\documentclass[conference]{IEEEtran}

\usepackage{cite}
\usepackage{amsmath,amssymb,amsfonts}
\usepackage{algorithmic}
\usepackage{graphicx}
\usepackage{textcomp}
\usepackage{xcolor}
\usepackage{booktabs}
\usepackage{multirow}
\usepackage{hyperref}
\hypersetup{colorlinks=false, pdfborder={0 0 0}}

\def\BibTeX{{\rm B\kern-.05em{\sc i\kern-.025em b}\kern-.08em
    T\kern-.1667em\lower.7ex\hbox{E}\kern-.125emX}}

\usepackage{microtype}
\begin{document}

\title{Beyond OCR Accuracy: Text-Centric VQA Under Image Degradation with Modular and End-to-End Pipelines}

\author{
\IEEEauthorblockN{Ritali Vatsi}
\IEEEauthorblockA{\textit{Indian Institute of Technology Mandi}\\
Mandi, India\\
ritali245994@gmail.com}
\and
\IEEEauthorblockN{Rachapudi Jagadeesh}
\IEEEauthorblockA{\textit{Indian Institute of Technology Mandi}\\
Mandi, India\\
S23096@students.iitmandi.ac.in}
\and
\IEEEauthorblockN{Shruti Singh Baghel}
\IEEEauthorblockA{\textit{Indian Institute of Technology Mandi}\\
Mandi, India\\
shrutibaghel19@gmail.com}
\linebreakand
\IEEEauthorblockN{Himani Sharma}
\IEEEauthorblockA{\textit{DIT University}\\
Dehradun, India\\
himanisharma781@gmail.com}
\and
\IEEEauthorblockN{Amit Shukla}
\IEEEauthorblockA{\textit{Indian Institute of Technology Mandi}\\
Mandi, India\\
amitshukla@iitmandi.ac.in}
\and
\IEEEauthorblockN{Pawan Goyal}
\IEEEauthorblockA{\textit{Indian Institute of Technology Kharagpur}\\
Kharagpur, India\\
pawang.iitk@gmail.com}
}

\makeatletter
\newcommand{\linebreakand}{%
  \end{@IEEEauthorhalign}
  \hfill\mbox{}\par
  \mbox{}\hfill\begin{@IEEEauthorhalign}
}
\makeatother

\maketitle

\begin{abstract}

Text-centric Visual Question Answering (VQA) requires reading and reasoning over text embedded in images, a task made substantially harder when images suffer from real-world degradation such as motion blur, low resolution, or compression artifacts. While modular OCR-based pipelines and end-to-end vision-language models are both widely used for this task, their comparative robustness under degraded conditions remains underexplored. We present an empirical study comparing two modular pipelines including SA-DBNet, a custom detection architecture combining ResNet-18 with self-attention spatial modeling and deformable convolutions against an end-to-end vision-language baseline, evaluated on 4013 degraded images with 7000 question-answer pairs. Fine-tuned modular pipelines achieve up to 57.50\% exact-match accuracy versus 38.00\% for the end-to-end baseline, with domain-specific fine-tuning yielding a gain of up to 29.50 percentage points. Critically, we find that conventional OCR error metrics like Character Error Rate and Word Error Rate are unreliable predictors of downstream VQA performance, as semantic reasoning can compensate for recognition failures when contextual cues are present. These findings highlight the importance of task-aware evaluation for text-centric VQA systems under realistic visual conditions. Codes are available  \href{https://github.com/RitaliVatsi/VQA_Project}{here}
\end{abstract}


\section{Introduction}

Visual Question Answering (VQA) seeks to respond to natural language inquiries on image content by concurrently analyzing visual and textual data. Recent advances in large language models have demonstrated increasingly capable reasoning and robustness across diverse language understanding tasks~\cite{rachapudi2026backflush, setti2026sedt, rachapudi2026bid, rachapudi2026repair}, raising the question of whether such capabilities can compensate for imperfect textual inputs in visual settings. In this larger task, ``text-centric VQA'' is a particularly challenging setting since answering a question requires correctly reading and understanding the text embedded in the image. In practice, such scenarios arise when reading product labels, deciphering signs, extracting numbers from printed materials, and analyzing language in documents or natural scenes. Text in real-world images is frequently degraded due to motion blur, low resolution, compression artifacts, or cluttered backgrounds. These issues often result in flawed optical character recognition (OCR), yielding noisy or partial textual inputs for downstream reasoning.

A widely held assumption in prior work is that higher OCR accuracy will automatically lead to better VQA performance. Recent advances in large multimodal models, however, suggest that semantic reasoning may in certain instances compensate for recognition failures, particularly when contextual cues are present. This raises an important and underexplored question: to what extent do OCR errors caused by visual degradation actually impair downstream VQA performance, and are conventional OCR error metrics reliable proxies for final answer quality?

\begin{figure}[t]
    \centering
    \includegraphics[width=1\columnwidth]{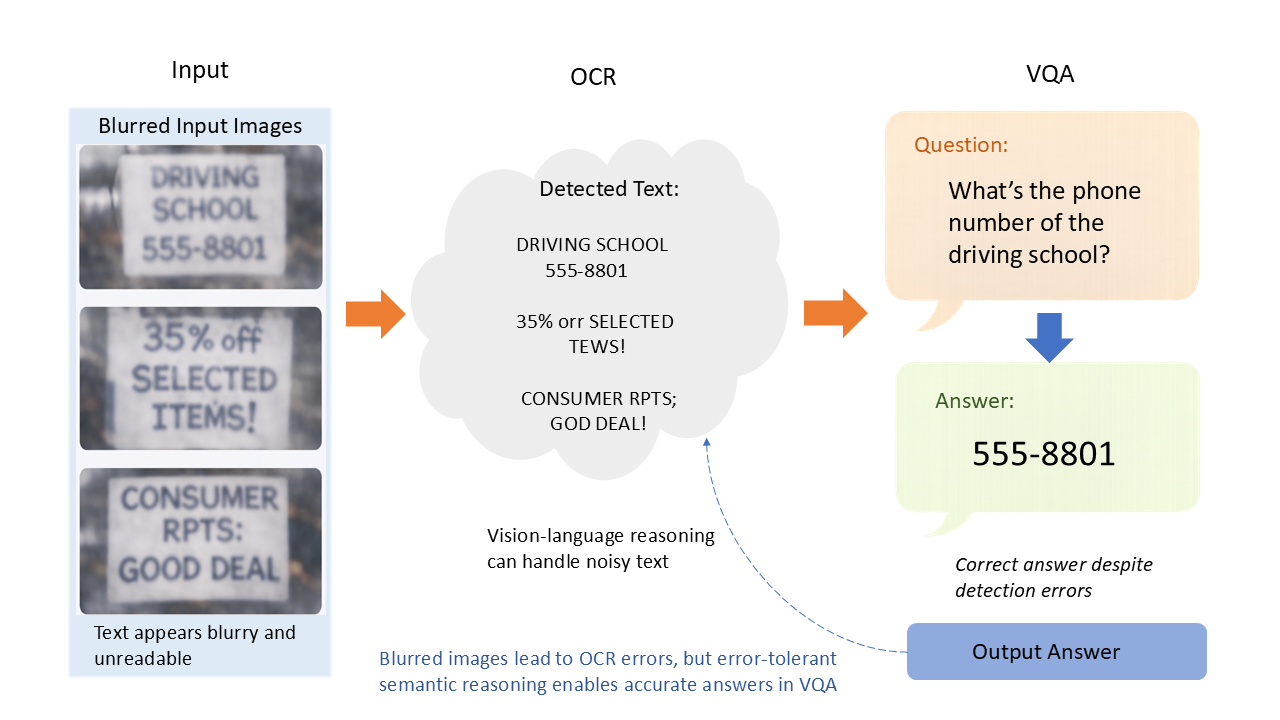}
    \caption{Conceptual illustration of text-centric VQA from a drone perspective. The system must recognize and reason over text (e.g., labeling) under motion blur degradation.}
    \label{fig:concept}
\end{figure}

In this study, we examine these questions through an empirical analysis of modular OCR-assisted pipelines and end-to-end vision-language models for text-centric VQA under compromised visual conditions. We construct a benchmark with 4013 images and 7000 question-answer pairs covering a variety of text types and layouts. Two modular pipelines---each comprising a text detection module, a text recognition module, and a language reasoning component---are evaluated against an end-to-end vision-language model. All systems are assessed using a standardized protocol covering exact correctness, string similarity, semantic alignment, and inference efficiency. The resulting comparison yields a systematic picture of how different architectural choices behave as visual quality degrades, and how OCR fidelity influences downstream answer quality.

This work makes three contributions. First, we present a benchmark framework for text-centric VQA under compromised visual conditions. Second, we conduct a controlled empirical comparison of modular OCR-assisted pipelines, their pretrained counterparts, and an end-to-end vision-language model, using both strict and robustness-focused evaluation criteria alongside runtime analysis. Third, we provide an empirical analysis of the relationship between OCR accuracy and final VQA performance, highlighting the role of downstream semantic reasoning in text-oriented VQA systems.

\section{Related Work}

Text-centric visual question answering (VQA) has been introduced to address scenarios where answering a question requires reading and reasoning over textual content embedded in images. Early benchmarks such as TextVQA~\cite{singh2019textvqa} and ST-VQA~\cite{biten2019stvqa} established the necessity of integrating optical character recognition (OCR) into VQA pipelines, demonstrating that conventional object-centric models fail when answers depend on scene text. Subsequent datasets, including OCR-VQA and DocVQA~\cite{mathew2021docvqa}, extended this paradigm to document images and structured layouts, further reinforcing the role of text extraction as a core component of visual reasoning.

In parallel, significant progress has been made in OCR itself. Robust text detectors such as CRAFT~\cite{baek2019craft} and DBNet~\cite{liao2020dbnet}, and recognizers such as TrOCR~\cite{li2023trocr} and SVTR~\cite{du2022ppocr}, have enabled more accurate transcription of irregular and degraded text. Many text-centric VQA systems explicitly leverage these advances by feeding recognized text tokens into multimodal reasoning models~\cite{hu2020m4c}, often assuming that improvements in OCR quality directly translate to improved VQA performance. Complementary work on scene text image restoration~\cite{wang2020textzoom} has similarly sought to improve downstream recognition by enhancing input image quality.

More recently, end-to-end vision-language models such as CLIP, Flamingo, BLIP-2, and Qwen-VL~\cite{bai2023qwen} have sought to bypass explicit OCR by jointly learning visual perception and semantic reasoning, suggesting that high-level context may compensate for imperfect text recognition. However, despite these developments, there is limited systematic analysis of how OCR errors under realistic visual degradation propagate to downstream VQA performance, or whether traditional OCR metrics such as CER and WER are reliable indicators of final answer correctness. Existing works primarily report aggregate VQA accuracy under standard conditions, leaving the relationship between low-level transcription noise and high-level semantic reasoning largely unexplored. This gap motivates our empirical study.

\section{Methodology}

Text-centric VQA requires jointly interpreting visual content and embedded text to answer natural language queries. In practical acquisition scenarios, textual cues are often observed under non-ideal visual conditions, including motion blur, defocus, and resolution loss. To explicitly capture this setting, let $I \in \mathbb{R}^{H \times W \times 3}$ denote an image and $q$ a corresponding question. The observed input is modeled as a degraded version of the original image, $\tilde{I} = \mathcal{D}(I)$, where $\mathcal{D}(\cdot)$ represents a visual degradation operator. The objective of text-centric VQA is therefore to predict an answer $a$ conditioned on the degraded visual input and the question:
\begin{equation}
a = f(\tilde{I}, q).
\end{equation}
This formulation makes explicit that all perception and reasoning stages must operate on imperfect visual evidence, which is central to the empirical analysis conducted in this work.

\subsection{Modular Pipeline Formulation}

A common paradigm for addressing text-centric VQA decomposes the mapping $f(\cdot)$ into modular stages corresponding to text detection, text recognition, and semantic reasoning. Under this formulation, the degraded image $\tilde{I}$ is first processed by a text detection function $r(\cdot)$ to produce a set of text regions:
\begin{equation}
B = r(\tilde{I}),
\end{equation}
where $B = \{b_1, b_2, \dots, b_K\}$ denotes the detected text bounding boxes. These regions are subsequently converted into textual tokens through a recognition function $h(\cdot)$:
\begin{equation}
T = h(B),
\end{equation}
yielding a sequence of recognized tokens $T = \{t_1, t_2, \dots, t_M\}$. Finally, a semantic reasoning module $g(\cdot)$ integrates the recognized text with the question to generate the predicted answer:
\begin{equation}
a = g(T, q).
\end{equation}
This modular decomposition enables the effects of OCR quality and semantic reasoning to be examined independently, allowing a controlled analysis of how recognition errors propagate to downstream VQA performance.

In contrast, end-to-end vision--language models directly infer answers from visual and linguistic inputs without explicitly separating OCR as an intermediate step:
\begin{equation}
a = f_{\text{VLM}}(\tilde{I}, q),
\end{equation}
where $f_{\text{VLM}}(\cdot)$ denotes a pretrained vision--language model. Such models implicitly learn joint visual--semantic representations and may therefore mitigate error amplification caused by explicit OCR failures. Including this formulation enables a direct comparison between modular and end-to-end approaches under identical degradation conditions.

\subsection{Evaluation Metrics}

To evaluate performance comprehensively, we employ metrics that capture both strict correctness and robustness to recognition noise. Exact-match accuracy measures the proportion of predictions that exactly match the ground-truth answers:
\begin{equation}
\text{Acc} = \frac{1}{N} \sum_{i=1}^{N} \mathbb{I}(\hat{a}_i = a_i),
\end{equation}
where $N$ denotes the total number of question--answer pairs. To account for partial correctness arising from minor transcription errors, we additionally report Average Normalized Levenshtein Similarity (ANLS):
\begin{equation}
\text{ANLS} = \frac{1}{N} \sum_{i=1}^{N} \max\left(0, 1 - \frac{D(a_i,\hat{a}_i)}{\max(|a_i|,|\hat{a}_i|)}\right),
\end{equation}
where $D(\cdot,\cdot)$ denotes the Levenshtein distance. Semantic similarity is computed as the cosine similarity between sentence embeddings produced by the \texttt{all-MiniLM-L6-v2} model from the \texttt{sentence-transformers} library, capturing meaning-level agreement beyond surface-form matching. For the end-to-end Qwen2-VL baseline, which does not produce an explicit OCR transcript, CER is computed by aligning the model's generated answer against the ground-truth answer text using standard edit-distance scoring; it is therefore reported as a diagnostic proxy rather than a direct measure of OCR quality. Character Error Rate (CER) and Word Error Rate (WER) are reported as diagnostic indicators of OCR quality, while Frames Per Second (FPS) quantifies end-to-end inference throughput.

\subsection{SA-DBNet: Self-Attention Enhanced Text Detection}

Existing text detectors such as DBNet~\cite{liao2020dbnet} perform well under clean imaging conditions but struggle with the sparse, geometrically distorted, and low-contrast text regions that characterize degraded imagery. To address these limitations, we introduce SA-DBNet, a purpose-built architecture designed for robust text localization under visual degradation. SA-DBNet extends the Differentiable Binarization (DB) framework~\cite{liao2020dbnet} with two key modifications targeted at the specific failure modes of standard detectors.

First, a self-attention spatial bottleneck is injected at the deepest backbone feature level. Operating on the 1/32-scale feature map from a ResNet-18~\cite{he2016resnet} backbone, the module computes query, key, and value projections through $1\times1$ convolutions, forming an attention map over all spatial positions. A learnable scaling parameter $\gamma$ controls the residual contribution of the attended features. This global spatial reasoning allows the detector to leverage long-range contextual dependencies, which is particularly valuable when text regions are small, sparse, or surrounded by visually similar clutter---conditions typical of degraded aerial imagery.

Second, the Feature Pyramid Network employs deformable convolutions (DCN-v2)~\cite{dai2017deformable} at every lateral output level. Unlike standard convolutions with fixed grid sampling, deformable convolutions learn data-dependent spatial offsets, enabling the receptive field to adaptively warp around geometrically distorted text. This is essential for handling perspective-skewed and curved text common in real-world captures.

SA-DBNet produces three output maps: a probability map $P$, a threshold map $T$, and a differentiable binary map $B = 1/(1 + e^{-k(P - T)})$ with $k = 50$. The model is trained using a composite loss combining Binary Cross-Entropy for the probability map, L1 loss for the threshold map, and Dice loss for the binary map. Training is performed using AdamW with gradient clipping. The model is first trained on TextVQA pseudo-labels (21,953 images) and then fine-tuned on ICDAR 2015 (1000 images) at $640\times640$ resolution, achieving an F1-score of 59\% on the detection benchmark. A component-level ablation isolating the individual contributions of the self-attention bottleneck and deformable convolutions relative to standard DBNet is identified as a direction for future work.

The first modular pipeline integrates SA-DBNet for text detection with transformer-based recognition via TrOCR~\cite{li2023trocr}, which formulates OCR as a sequence-to-sequence translation task using a Vision Transformer encoder and an autoregressive language decoder.

\begin{figure}[t]
\includegraphics[width=1\linewidth]{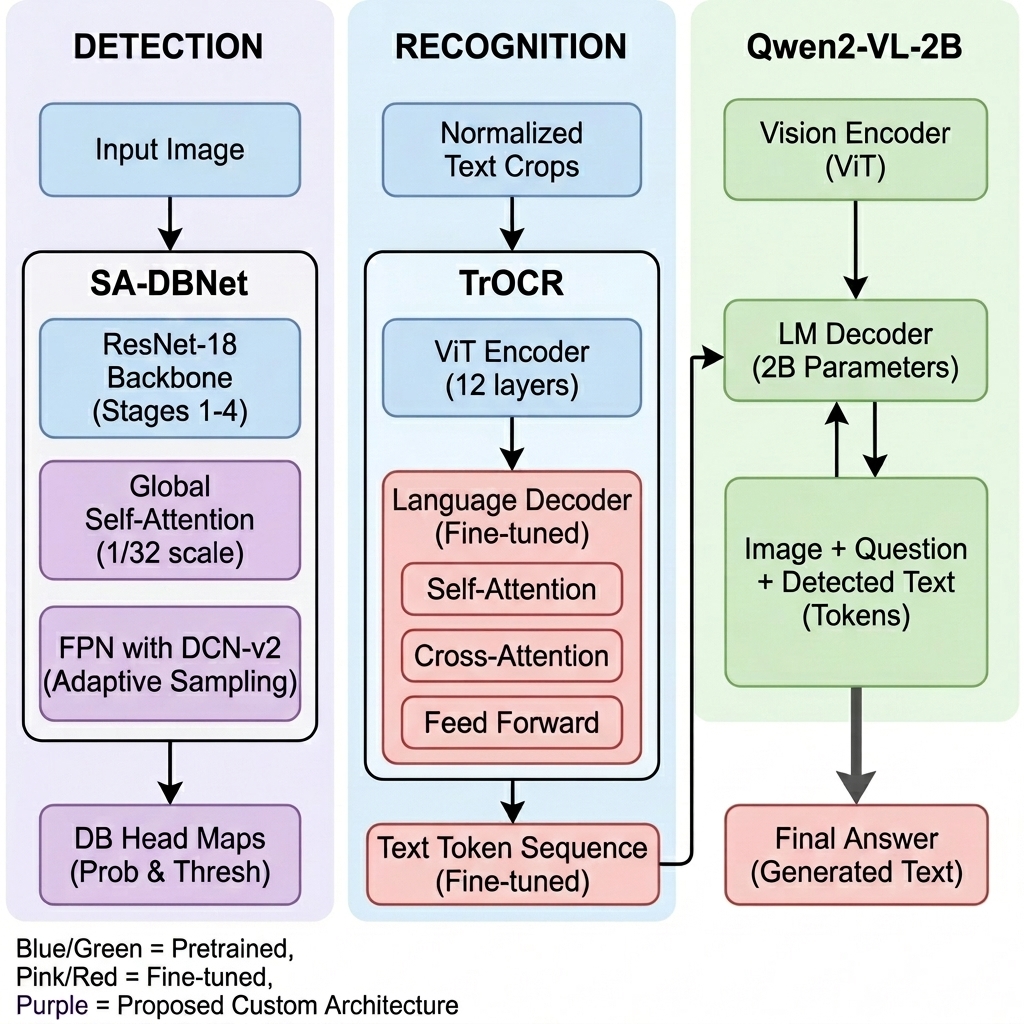}
\caption{Proposed modular pipeline: SA-DBNet + TrOCR + Qwen2-VL. Color coding: blue = pretrained components, purple = proposed custom architecture (self-attention bottleneck, DCN-v2 FPN, DB head), pink/red = fine-tuned components, green = pretrained reasoning module.}
\label{fig:sadbnet_pipeline}
\end{figure}

\subsection{PaddleOCR-Based Pipeline}

The second modular pipeline is designed to emphasize inference efficiency and practical deployability. It is based on PP-OCRv4~\cite{du2022ppocr}, which employs DB++ with a MobileNetV3 backbone for text detection, coupled with a DBFPN neck and a DBHead for differentiable binarization. Text recognition is performed using SVTR, which balances representational capacity and computational efficiency through spatial mixing operations. This pipeline serves as a strong, lightweight baseline against which the SA-DBNet pipeline is compared.

\begin{figure}[t]
\includegraphics[width=1\linewidth]{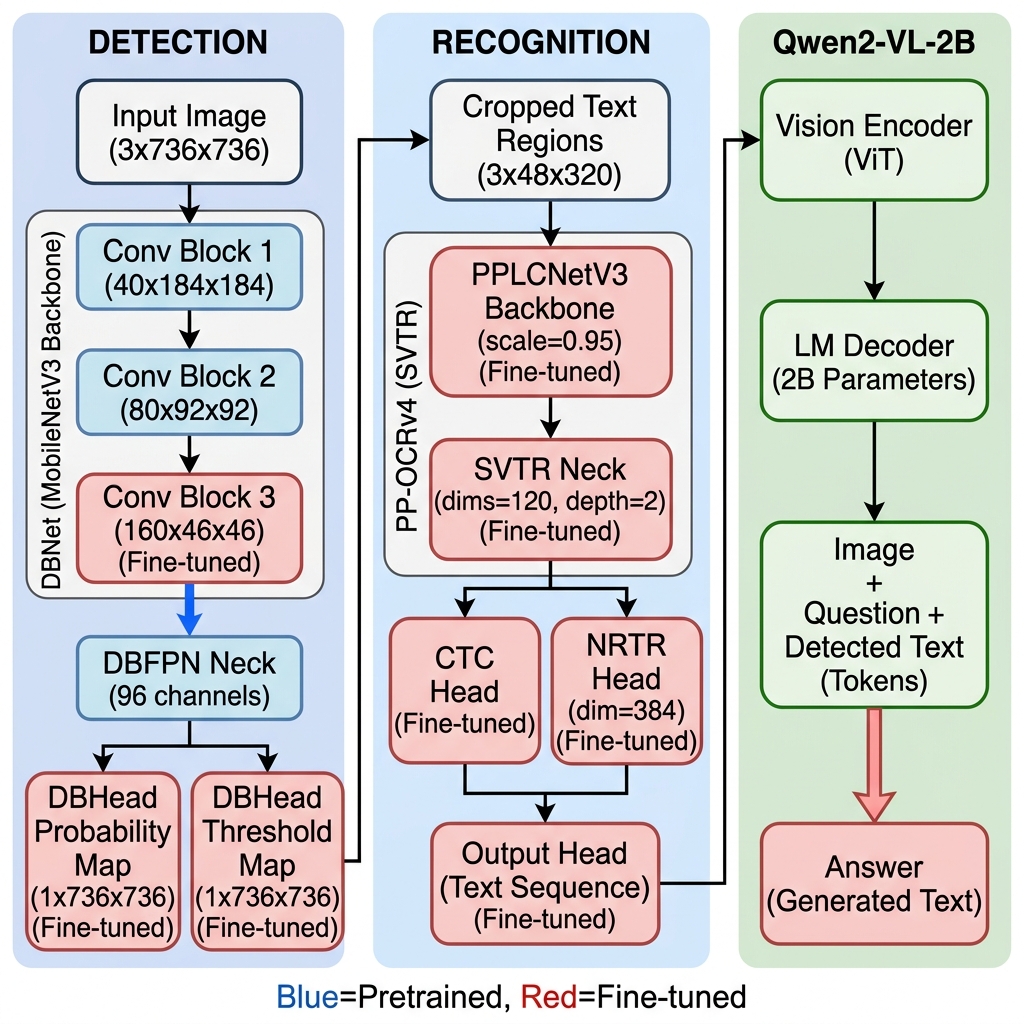}
\caption{Baseline PaddleOCR + Qwen pipeline architecture showing the standard DBNet+SVTR modular approach for benchmark comparison.}
\label{fig:paddleocr_pipeline}
\end{figure}

\subsection{Shared Reasoning Component and Fine-Tuning Strategy}

Both modular pipelines share the same reasoning component: Qwen2-VL-2B, ensuring that observed differences in downstream VQA performance are primarily attributable to OCR quality rather than reasoning capacity. During inference, Qwen2-VL-2B operates with 4-bit quantization and Flash Attention~2 to reduce memory usage and improve throughput.

To enhance robustness under severe visual degradation, targeted fine-tuning strategies are applied to the OCR components. PaddleOCR is trained using a three-stage curriculum learning scheme~\cite{bengio2009curriculum} with progressively increasing blur severity. TrOCR is fine-tuned on blur-augmented data starting from pretrained \texttt{trocr-base-stage1} weights, incorporating Gaussian blur, motion blur, and resolution degradation during training while keeping the Vision Transformer encoder frozen.

\section{Experiments and Results}

This section presents a detailed empirical evaluation of the two modular pipelines and the end-to-end vision--language model under degraded visual conditions. We first describe the benchmark and experimental protocol, then present a comparative performance analysis, an assessment of domain-specific fine-tuning, and a focused examination of OCR quality metrics in relation to downstream VQA accuracy.

\subsection{Dataset and Experimental Setup}

The evaluation is conducted on a curated benchmark consisting of 4013 images associated with a total of 7000 question--answer pairs. The images are sourced from publicly available text-centric scene datasets, covering real-world environments such as storefronts, street signs, product packaging, and printed documents. Question--answer pairs were constructed through a semi-automated process in which candidate questions were generated based on visible text regions in each image and subsequently verified manually to ensure answer correctness and question clarity. The dataset encompasses a wide range of textual content, visual layouts, and scene contexts. Images exhibit mixed viewpoints, background clutter, and non-uniform illumination, all of which pose challenges for both text extraction and semantic reasoning.

To simulate acquisition artifacts commonly encountered in mobile and aerial imaging systems, visual degradation is applied uniformly across the dataset using Gaussian blur with standard deviation $\sigma = 5.0$. This is the primary experimental condition used for all pipeline comparisons, ensuring consistency and reproducibility across all systems. In addition to Gaussian blur, we consider representative examples of motion blur, median filtering, and downscaling to illustrate the diversity of degradation types that may arise in practice, as illustrated in Fig.~\ref{fig:degradation_examples}; a systematic evaluation across multiple degradation types is identified as an important direction for future work.

\begin{figure}[t]
\centering
\includegraphics[width=\columnwidth]{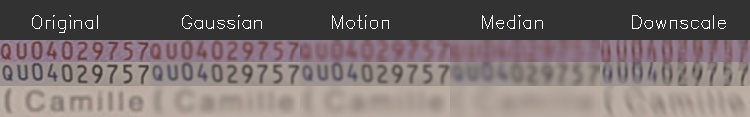}
\caption{Examples of visual degradation applied to text images in the experimental setup, including Gaussian blur, motion blur, median filtering, and downscaling.}
\label{fig:degradation_examples}
\end{figure}

Questions are categorized into three groups based on their reasoning requirements: (i) header-level queries that depend on prominent textual cues, (ii) detail-level queries requiring accurate recognition of fine-grained or numerical text, and (iii) spatial reasoning queries involving the relative arrangement of textual elements. This categorization enables a comprehensive evaluation of OCR fidelity and semantic reasoning under degraded visual input. Disaggregated results by question category are left for future analysis.

\subsection{Implementation Details}

For modular pipelines, text detection and recognition are implemented using established open-source frameworks. The SA-DBNet-based pipeline employs the custom SA-DBNet detector for text localization and a fine-tuned TrOCR model for recognition, while the PaddleOCR-based pipeline uses PP-OCRv4 with curriculum-trained weights optimized for blur robustness. In both cases, recognized text tokens are provided to Qwen2-VL-2B through a unified prompting interface, ensuring that differences in performance arise primarily from OCR quality rather than variations in reasoning input.

The end-to-end baseline directly applies the Qwen2-VL model to the degraded image and question without explicit OCR decomposition. All experiments are conducted under identical hardware and inference settings. End-to-end throughput is reported in Frames Per Second (FPS), accounting for the complete inference pipeline, including preprocessing, OCR (when applicable), and semantic reasoning.

\subsection{Comparative Performance Analysis}

Table~\ref{tab:main_results} presents the primary quantitative comparison under heavy visual degradation ($\sigma = 5.0$). Both fine-tuned modular pipelines demonstrate superior exact-match accuracy compared to the end-to-end baseline. The SA-DBNet+TrOCR pipeline achieves the highest accuracy at 57.50\%, followed by PaddleOCR at 42.83\%. In comparison, the end-to-end Qwen2-VL model attains 38.00\%. These results suggest that explicit text extraction provides stronger grounding for VQA when inputs are degraded.

However, the end-to-end model achieves comparable throughput at 0.76 FPS versus 0.72 FPS for PaddleOCR+Qwen---a marginal 5.6\% difference---while SA-DBNet+TrOCR operates at 0.59 FPS due to its more heavyweight recognition stage. Notably, the end-to-end model leads on semantic similarity (0.65), suggesting it produces answers meaningfully closer to ground truth even when they are not exact matches.

A complementary observation is that SA-DBNet+TrOCR achieves the highest exact-match accuracy yet records lower ANLS (0.54 vs.\ 0.59) and semantic similarity (0.50 vs.\ 0.61) than PaddleOCR. This pattern suggests that SA-DBNet+TrOCR tends to produce answers that are either fully correct or substantially different from the ground truth, whereas PaddleOCR produces more consistently near-correct responses. This reflects a difference in recognition behavior---precision-oriented versus recall-oriented---and underscores why both exact-match and similarity-based metrics are necessary for a complete evaluation.

\begin{table}[t]
\centering
\caption{Main Performance Comparison (Blur $\sigma=5.0$, $n=7000$)}
\label{tab:main_results}
\setlength{\tabcolsep}{4pt}
\renewcommand{\arraystretch}{1.08}
\scriptsize
\resizebox{\columnwidth}{!}{%
\begin{tabular}{lccccc}
\toprule
Pipeline & Acc (\%) & ANLS & Semantic & CER & FPS \\
\midrule
\textbf{PaddleOCR+Qwen (FT)}      & 42.83          & \textbf{0.59} & 0.61          & \textbf{0.42} & 0.72          \\
\textbf{SA-DBNet+TrOCR+Qwen (FT)} & \textbf{57.50} & 0.54          & 0.50          & 0.46          & 0.59          \\
Qwen2-VL (End-to-End)             & 38.00          & 0.57          & \textbf{0.65} & 0.58          & \textbf{0.76} \\
\bottomrule
\end{tabular}%
}
\end{table}

\subsection{Impact of Domain-Specific Fine-Tuning}

To isolate the effect of OCR adaptation, we compare fine-tuned pipelines against their pretrained counterparts under identical blur conditions. Results are summarized in Table~\ref{tab:finetuning}. Fine-tuning yields substantial improvements for both pipelines: PaddleOCR improves from 32.50\% to 42.83\%, while SA-DBNet+TrOCR improves from 28.00\% to 57.50\%, a gain of 29.50 percentage points.

These results suggest that domain-specific exposure to degraded visual inputs is important for robust text recognition in text-centric VQA. The larger relative gain observed for SA-DBNet+TrOCR highlights the adaptability of transformer-based recognition models when trained under challenging visual conditions, suggesting that the architecture has greater headroom for task-specific specialization.

\begin{table}[t]
\centering
\caption{Impact of Fine-Tuning on Blur Robustness}
\label{tab:finetuning}
\begin{tabular}{lccc}
\toprule
Pipeline & Pretrained & Fine-tuned & Gain \\
\midrule
PaddleOCR+Qwen               & 32.50\% & 42.83\%          & +10.33\% \\
\textbf{SA-DBNet+TrOCR+Qwen} & 28.00\% & \textbf{57.50\%} & \textbf{+29.50\%} \\
\bottomrule
\end{tabular}
\end{table}

\subsection{OCR Quality and Downstream VQA Performance}
\label{sec:ocrquality}

Table~\ref{tab:ocr_quality} reports OCR-specific quality metrics for the fine-tuned pipelines. PaddleOCR achieves a lower Character Error Rate (0.42) compared to SA-DBNet+TrOCR (0.46), yet scores substantially lower in downstream VQA accuracy (42.83\% vs.\ 57.50\%). This inversion illustrates a key finding: lower aggregate CER does not guarantee better VQA performance. The nature of the recognition errors matters more than their raw frequency---missing a critical keyword such as a product name or numerical value is far more damaging to downstream reasoning than minor character-level substitutions in peripheral text.

\begin{table}[t]
\centering
\caption{OCR Quality Metrics (Fine-Tuned Models)}
\label{tab:ocr_quality}
\begin{tabular}{lccc}
\toprule
Pipeline & CER $\downarrow$ & WER $\downarrow$ & Semantic $\uparrow$ \\
\midrule
PaddleOCR (FT)               & \textbf{0.42} & \textbf{0.50} & \textbf{0.61} \\
\textbf{SA-DBNet+TrOCR (FT)} & 0.46          & 0.56          & 0.50          \\
\bottomrule
\end{tabular}
\end{table}

\section{Ablation Study}

To better understand the factors driving text-centric VQA performance under visual degradation, we conducted ablation experiments isolating the contributions of explicit OCR integration, the text detection stage, and the relationship between OCR error metrics and downstream reasoning. All ablations are performed under identical blur conditions ($\sigma = 5.0$).

\subsection{Effect of Explicit OCR Integration}

Table~\ref{tab:ablation_explicit} extends the main comparison by highlighting the trade-offs between modular and end-to-end designs. Including explicit OCR consistently improves exact-match accuracy---the best modular system (SA-DBNet+TrOCR+Qwen) outperforms the end-to-end model by 19.5 percentage points. However, the end-to-end model retains advantages in semantic similarity (0.65) and inference throughput (0.76 FPS), indicating that the choice of paradigm should be guided by application requirements: strict textual correctness vs.\ efficiency and soft semantic alignment.

\begin{table}[h]
\centering
\caption{Modular vs.\ End-to-End Approaches (Blur $\sigma=5.0$)}
\label{tab:ablation_explicit}
\begin{tabular}{lccc}
\toprule
\textbf{Pipeline Design} & \textbf{Accuracy} & \textbf{Sem. Sim.} & \textbf{FPS} \\
\midrule
End-to-End (Qwen2-VL)                  & 38.00\%          & \textbf{0.65} & \textbf{0.76} \\
Modular (PaddleOCR+Qwen)               & 42.83\%          & 0.61          & 0.72          \\
\textbf{Modular (SA-DBNet+TrOCR+Qwen)} & \textbf{57.50\%} & 0.50          & 0.59          \\
\bottomrule
\end{tabular}
\end{table}

\subsection{Importance of Text Detection}

To assess the necessity of a dedicated text detector, we evaluated the SA-DBNet+TrOCR pipeline with and without the detection stage, feeding the full scene image directly to the recognizer in the latter case. Isolated evaluation of SA-DBNet confirms its standalone detection capability with an F1-score of 59\%. As shown in Table~\ref{tab:ablation_detection}, removing the detection stage causes a severe performance collapse (ANLS $0.54 \rightarrow 0.09$), confirming that even powerful sequence recognizers like TrOCR require spatially localized input to function correctly in complex, cluttered scenes.

\begin{table}[h]
\centering
\caption{Ablation of Text Detection Module (ANLS)}
\label{tab:ablation_detection}
\begin{tabular}{lc}
\toprule
\textbf{Configuration} & \textbf{ANLS Score} \\
\midrule
Full Pipeline (Det + Rec) & \textbf{0.54} \\
No Detection (Rec Only)   & 0.0886        \\
\bottomrule
\end{tabular}
\end{table}

\subsection{Relationship Between OCR Quality and VQA Accuracy}

The CER/WER inversion observed in Section~\ref{sec:ocrquality} is further examined here. The end-to-end model, which produces no explicit OCR transcript and thus has the highest effective recognition error, still answers 38\% of questions correctly by relying on contextual semantic reasoning. Meanwhile, PaddleOCR achieves the best CER (0.42) among the modular systems yet lags behind SA-DBNet+TrOCR by 14.67 percentage points in VQA accuracy. Taken together, these results suggest that CER and WER are insufficient standalone predictors of VQA success. Task-specific metrics such as ANLS, which reward partial credit for near-correct answers, better reflect the semantic tolerance of downstream reasoning modules.

\section{Discussion}

The experimental results reveal several important properties of text-centric VQA systems under degraded visual conditions. Domain-specific fine-tuning of OCR components substantially improves downstream VQA performance. Curriculum-based blur training and blur-aware augmentation enable OCR models to handle challenging visual input without sacrificing general capability, confirming that alignment between training and deployment distributions is a practical strategy. The comparison between modular and end-to-end paradigms reveals complementary strengths. Modular pipelines achieve higher exact-match accuracy under severe degradation through explicit textual grounding, while the end-to-end model offers higher throughput and competitive semantic similarity, suited for efficiency-oriented scenarios. Neither paradigm strictly dominates; system selection should reflect the accuracy-efficiency trade-off of the target application. A key finding is that OCR quality and VQA performance are not tightly coupled. Even with significant recognition errors, semantic reasoning modules frequently recover correct answers via contextual and linguistic cues. The \emph{type} of error matters more than the aggregate rate: mistakes on answer-critical tokens---such as proper nouns, numeric values, or identifiers---disproportionately harm accuracy, while errors on peripheral text are tolerated. This implies that metrics such as CER and WER should be supplemented by task-aware error weighting distinguishing answer-relevant from non-answer-relevant text. These findings reinforce the value of holistic, system-level evaluation for text-centric VQA.

\section{Conclusion}
This paper presented an empirical comparison of modular OCR-assisted pipelines and an end-to-end vision-language model for text-centric VQA under degraded visual conditions. Fine-tuned modular pipelines achieve higher exact-match accuracy through explicit textual grounding, while the end-to-end model offers competitive similarity-based robustness with higher throughput. The analysis shows that conventional OCR metrics such as CER and WER do not consistently reflect downstream VQA performance, as semantic reasoning can compensate for recognition failures via contextual cues. This underscores the need for holistic, task-aware evaluation of text-centric VQA systems. Future work will extend evaluation to additional degradation types, disaggregate results by question category, expand the dataset, conduct component-level ablation of SA-DBNet, and explore hybrid architectures integrating explicit OCR with end-to-end reasoning.


\end{document}